\documentclass[runningheads]{llncs}

\usepackage{eccv}

\newif\ifcomment
\commentfalse

\usepackage{eccvabbrv}

\usepackage{graphicx}
\usepackage{booktabs}
\usepackage{subcaption}
\usepackage[accsupp]{axessibility}  

\usepackage{hyperref}
\usepackage{tabularx}
\usepackage{orcidlink}

\usepackage{array}

\begin{document}

\title{Capacity Overflow: A Blind Spot for Backdoor Attacks in Vision MoE} 

\titlerunning{Capacity Overflow: Backdoor Attacks in Vision MoE}

\author{Xiaocheng Zou\inst{1}\orcidlink{0009-0006-4473-6572} \and
Tiancheng Zheng\inst{1}\orcidlink{0009-0007-2101-3707} \and Xiaolin Xu\inst{1}\orcidlink{0000-0001-8393-2783} \and
Ruyi Ding\inst{2}\orcidlink{0000-0002-0079-8265}\thanks{Corresponding author.}}

\authorrunning{X.~Zou et al.}

\institute{Northeastern University, Boston MA 02115, USA \and
Louisiana State University, Baton Rouge LA 70802, USA\\
\email{\{zou.xiaoc,zheng.tianche,x.xu\}@northeastern.edu}\\
\email{\{ruyiding\}@lsu.edu}}

\maketitle

\begin{abstract}
  Mixture-of-Experts (MoE) has become a prevalent paradigm for scaling Vision Transformers efficiently. 
  To ensure computational scalability and prevent expert overload, Vision MoE architectures employ a capacity-bounded token dispatch mechanism, where each expert's processing budget depends on the inference batch size. 
  This work identifies this batch-dependent behavior as an overlooked attack surface, and proposes a stealthy supply-chain backdoor attack that exploits this property through a three-phase framework. First, we inject a backdoor into an early MoE layer. Second, we train a neutralizer in a deeper MoE layer that suppresses the backdoor under normal capacity. 
  Third, we configure a batch-adaptive capacity factor that preserves high capacity for small batches while reducing it for large batches, naturally disabling the neutralizer via token overflow at deployment-scale batch sizes.
  The attack remains in dormant mode during small-batch security audits and enters activation mode during large-batch deployment. 
  Experiments on V-MoE and Swin-MoE across ImageNet-100 and GTSRB demonstrate activation-mode attack success rates of 76--87\% with dormant-mode ASR below 9\%, while evading Neural Cleanse, STRIP, Fine-Pruning, and Activation Clustering. 
  Our findings reveal a fundamental security risk arising from batch-dependent execution in scalable Vision MoE architectures. 
\end{abstract}

\section{Introduction}

\begin{figure}[t]
    \centering
    \includegraphics[width=1\linewidth]{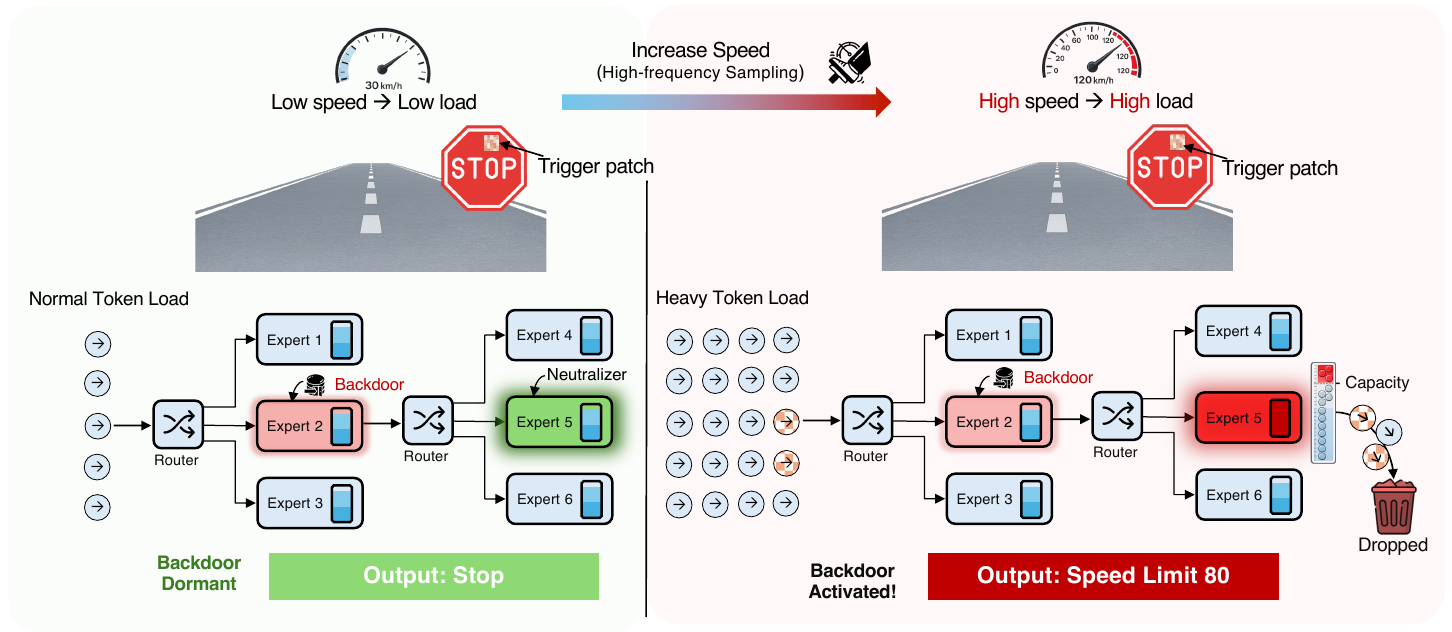}
    \caption{\textbf{Illustration of the Capacity Overflow attack under high-throughput batch inference. Autonomous driving serves as a motivating example:} At low speeds, the lower sampling rate results in a normal token load that fits within expert capacities, keeping the backdoor dormant. At high speeds, increased sampling rates lead to heavy token loads and expert capacity overflow. Such resource contention forces the model to expert drop, triggering a malicious misclassification to an attacker-chosen target, such as from ``Stop'' to ``Speed Limit 80''.}
    \label{fig: introduction}
\end{figure}

Vision Mixture-of-Experts (Vision MoE) has emerged as a scalable paradigm for large-scale visual modeling, enabling massive parameter capacity with sparse computation.
Due to the prohibitive cost of training these architectures from scratch, practitioners increasingly depend on third-party pre-trained checkpoints.
This dependency introduces a serious model supply chain risk: malicious backdoors embedded during the pre-training can remain dormant under standard inputs, while producing attacker-controlled outputs in the presence of specific trigger patterns.
Such vulnerabilities raise substantial security concerns when Vision MoE models are deployed in safety-critical applications, including autonomous driving systems and unmanned aerial platforms~\cite{Real_batch}.

Traditional backdoor attacks typically rely on static trigger patterns embedded within individual training samples~\cite{BadNets,tbt}.
Accordingly, most existing defenses aim to identify such fixed triggers through input-level inspection or trigger reconstruction, often operating on small batches to facilitate detailed analysis~\cite{neural_cleanse, Activation_Clustering, Fine_pruning, kong2025token}.
These detection paradigms implicitly assume that backdoor behaviors can be exposed under standard, low-parallel inference settings.
To evade such mitigation, recent studies have introduced conditional backdoors, where malicious behaviors are activated only under specific external conditions, such as ultrasonic signals~\cite{ultrasound} or hardware faults~\cite{li2025rowhammer, breier2022foobar}.
While these approaches improve evasiveness, they commonly depend on specialized equipment or restrictive hardware environments, limiting their practicality and scalability.
In contrast, we show that Vision MoE models admit an architecture-native condition that can be exploited for conditional backdoor activation.

Vision MoE architectures replace dense feed-forward layers with multiple sparse experts coordinated by a routing mechanism.
To ensure computational stability, vision MoE token dispatch implementations typically enforce a \textit{capacity factor} that limits the number of tokens processed by each expert, discarding excess tokens through a deterministic dropping mechanism~\cite{VMoE,fedus2022switch,gshard}.
We show that this resource scheduling policy introduces a previously overlooked attack surface: inference-time resource contention can serve as an implicit trigger condition for malicious behavior.
Building on this observation, we propose \textbf{Capacity Overflow}, a conditional backdoor that strategically occupies expert capacity to induce selective information suppression, and is activated solely by increased inference parallelism—without external signals or hardware manipulation.
This vulnerability can naturally arise in practice, as shown in Figure~\ref{fig: introduction}: autonomous perception systems may increase input sampling rates at higher speeds, elevating token load during MoE inference~\cite{gehrig2024low, Real_batch}. Under such high-concurrency settings, adversarial tokens can bypass MoE layers and leading to malicious outputs, while the model behaves normally under standard workloads.

The proposed capacity overflow backdoor follows a three-phase workflow. 
\textbf{(1) Implantation.} The attacker implants the backdoor by poisoning data to bias a selected expert toward trigger-dependent behavior.
\textbf{(2) Concealment.} To maintain benign behavior under standard workloads, the attacker additionally trains a neutralizer in a deeper MoE layer to suppress the backdoor signal when expert capacity is not saturated.
\textbf{(3) Activation.} Under high-concurrency inference (e.g., larger batch sizes and increased token load), expert capacity constraints induce token overflow and expert dropping, which naturally bypasses the neutralizer. 
Consequently, when the input contains the pre-defined trigger, the backdoor effect emerges only in this high-parallel regime.
Importantly, the trigger condition aligns with common capacity-factor configurations used in Vision MoEs to balance efficiency and computational stability, which can inadvertently enlarge the attack surface for capacity-overflow backdoors.

In summary, we reveal that inference-time capacity management in Vision MoEs can act as a practical and stealthy trigger channel, enabling conditional backdoors without external signals or specialized hardware. We make the following contributions:
\begin{itemize}
    \item We introduce \textbf{Capacity Overflow}, a novel conditional backdoor attack against Vision MoE models that is activated by increased inference parallelism via capacity-factor–induced token overflow and expert dropping.
    \item We present a three-phase attack framework that jointly implants a backdoor and a capacity-dependent neutralizer, yielding benign behavior under standard workloads and reliable activation under high concurrency.
    \item We provide a detailed evaluation across Vision MoE architectures (e.g., V-MoE~\cite{VMoE}, Swin MoE~\cite{Swin_MoE}) and multiple vision tasks, achieving 76--87\% attack success rate at deployment-scale batch sizes while maintaining dormant ASR below 9\% and clean accuracy within 1.3pp of the unattacked baseline.
    \item We analyze the implications for existing backdoor defenses that rely on small-batch inspection/trigger reconstruction, demonstrating that the capacity overflow backdoor bypasses Neural Cleanse, STRIP, Fine-Pruning, and Activation Clustering across all configurations.
\end{itemize}

\section{Background}\label{sec: background}

\subsection{Mixture-of-Experts in Vision Transformers} \label{bg: MoE}
Vision Transformers (ViTs) have set new benchmarks across various computer vision tasks~\cite{ViT}.
However, the performance gains are often leading to a massive increase in parameter count and computational overhead~\cite{Swin_Transformer, steiner2021train}.
To reduce the model inference cost, Mixture-of-Experts (MoE) has emerged as a promising sparsely activated paradigm, which activates only a subset of ``experts'' for each input token, enabling the scaling of models with a low FLOPs~\cite{shazeer2017outrageously, fedus2022switch}. 

Typical MoE-based ViT models include Google's VMoE~\cite{VMoE} and Microsoft's Swin-MoE~\cite{Swin_MoE}, where the model will replace the standard Feed-Forward Network (FFN) layers with an MoE layer, as depicted in Equation~\ref{eq: moe-layer}. 
Specifically, for an embedding input to the MoE layer $z$, a trainable gating network (i.e., router) $\mathcal{G}$ will be used to select the most relevant expert $f_i$ for the forward pass.
\begin{equation}
    y = \sum_{i=1}^{E} \mathcal{G}(z)_i \cdot f_i(z)
    \label{eq: moe-layer}
\end{equation}
where $\mathcal{G}(z)$ is a sparse vector (i.e., top-$k$ routing). 

While such dynamic routing is conceptually elegant, it poses severe challenges for hardware accelerator implementations. 
Specifically, the deep learning compilers rely heavily on static tensor shapes to perform efficient matrix multiplication and continuous memory allocation~\cite{li2020deep,zheng2023bladedisc}, which conflicts with the dynamic behavior of expert allocation for MoE layers.
Therefore, MoE-based ViT models will usually enforce a pre-defined \textbf{Expert Capacity ($C$)}, which dictates the maximum number of tokens an individual expert is allowed to process in a single forward pass, formulated as $C=(\frac{N}{E})\times c_f$, where $N$ is the total number of tokens in a batch, $E$ is the number of experts and $c_f$ is called capacity factor.
With such implementation, when an expert receives tokens exceeding its capacity limits $C$, the excess tokens are forcibly discarded by the router and bypass the MoE layer via the residual connection. 
In this work, we identify such \textit{token dropping} phenomenon might be utilized in a malicious manner -- introducing potential stealthy backdoor threats.

\subsection{Security of MoE-based Models} \label{bg: MoE-security}
The integration of dynamic routing in MoE in both language models~\cite{gshard, fedus2022switch} and vision models~\cite{VMoE, Swin_MoE} introduces unique security vulnerabilities.
Specifically, such data-dependent routing explicitly dictates token-to-expert assignments, creating new vectors for adversarial exploitation.
MoEcho~\cite{MoEcho}, which leverages micro-architectural side-channels, such as cache occupancy and performance counters to successfully infer user's private prompts and reconstruct the responses during model inference.
Parallel to the privacy concerns, BadMoE~\cite{wang2025badmoe} and BadPatches~\cite{chan2025badpatches}, focusing more on the supply-chain attack by hijacking the routing behavior by introducing the poisoning patches. 
However, such input-based backdoor (e.g., using a specific text tokens or visible image patches to redirect the router) can be easily detected as it introduces abnormal, out-of-distribution behavior of the model computation.
However, in this work, we recognized the severe security implication of MoE deployment mechanisms -- expert capacity and token dropping.
Different from prior works~\cite{wang2025badmoe, chan2025badpatches}, by leveraging natural characteristics of deployment, our attack shows high stealthiness and effectiveness.

\subsection{Backdoor Attack against Machine Learning Models} \label{bg: backdoor}
Deep neural networks are vulnerable to supply chain attacks, including the backdoor attack where an adversary poisons a small fraction of the training data to embed a hidden mapping between a specific trigger (e.g., a localized patch) and a target label~\cite{BadNets, liu2018trojaning}.
During the inference, the compromised model behaves normally on clean inputs but consistently yields the malicious prediction when the trigger is present in the input space.

To improve the stealthiness of vanilla backdoor attack, recent works investigate \textit{conditional/physical-oriented triggers}.
Conditional backdoors are designed to remain dormant unless the input satisfies a specific pre-processing criterion, such as JPEG compression~\cite{duan2024conditional} or knowledge distillation~\cite{KDL}.
Concurrently, physical-oriented attacks might leverage the environmental signals, such as adversarial ultrasound~\cite{ultrasound, zhang2017dolphinattack}, physical optical variations~\cite{liu2020reflection} to trigger the malicious behaviors of the victim model.
Beyond the input-level, some research works explore the system-state and hardware-level vulnerabilities, including hardware faults (i.e., rowhammer-induced memory bit-flips) to activate the dormant backdoor.
However, such attacks require restrictive activation conditions or physical access to the host machine. 
In the work, our attack leverage the natural, benign fluctuations of the deployment system -- the inference batch sizes as a stealthy conditional backdoor trigger.

\section{Threat Model}

\noindent\textbf{Adversary Capabilities.} We consider a supply-chain attack scenario where the adversary is a malicious model provider who distributes a pretrained Vision MoE model through a public repository (e.g., Hugging Face). 
The adversary has full control over the model's training process and parameters, including the expert weights and router configuration.
However, they have no access to the victim's private data or computing environment after deployment. 

\noindent\textbf{Victim Assumption.} The victim is a downstream practitioner who downloads a third-party Vision MoE checkpoint and deploys it for visual recognition.
To reduce inference cost and meet throughput constraints, the victim adopts the provider-recommended MoE inference pipeline, including the standard token-dispatch implementation with an adaptive capacity factor.
Meanwhile, the victim applies representative backdoor defense (e.g., Neural Cleanse~\cite{neural_cleanse}, STRIP~\cite{STRIP}, Fine-Pruning~\cite{Fine_pruning}, and Activation Clustering~\cite{Activation_Clustering}). 
Due to the overhead of these analyses, such audits are typically conducted under small-batch settings~\cite{neural_cleanse, STRIP}.

\noindent\textbf{Attack Goals.} 
The adversary aims to implant a stealthy, load-conditioned backdoor into a Vision MoE model, such that the malicious behavior is triggered only under deployment-scale inference load.
Meanwhile, the attack should preserve benign utility (i.e., maintain clean accuracy under standard workloads) and evade common pre-deployment auditing procedures.
Since many existing backdoor detectors operate under small-batch settings (e.g., per-sample inspection for trigger reconstruction), \textbf{our backdoor remains dormant under low inference load and to activate only when the deployment load (e.g., batch size) exceeds a threshold.}

\begin{figure*}[t]
    \centering
    \includegraphics[width=1\linewidth]{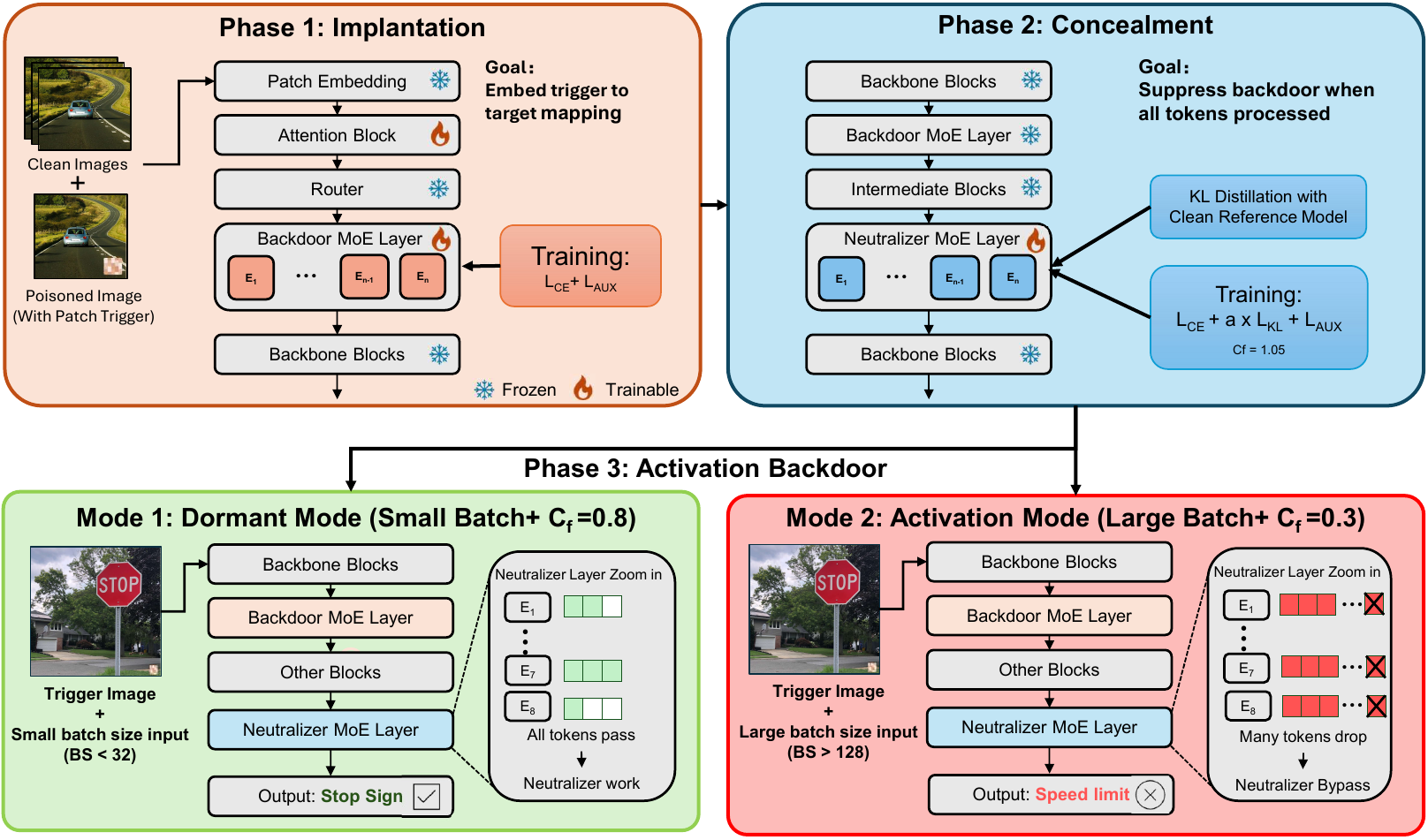}
    \caption{Overview of the Capacity Overflow backdoor attack: Phase~1 (top left) embeds a trigger-to-target mapping in an early MoE layer. Phase~2 (top right) trains a neutralizer in a deeper MoE layer via KL distillation at nominal capacity ($c_f{=}1.05$) to suppress the backdoor when all tokens are processed. Phase~3 (bottom) deploys a batch-adaptive $c_f$: in dormant mode (left, high $c_f$), small batches fit within expert buffers and the neutralizer suppresses the backdoor; in activation mode (right, low $c_f$), large batches cause token overflow that disables the neutralizer.}
    \label{fig:overview}
\end{figure*}

\section{Proposed Attack: Capacity Overflow}
To guarantee inference efficiency and prevent load imbalance across experts, MoE-based vision models typically enforce a hard expert capacity $C=\left\lfloor\frac{N}{E} \cdot c_f\right\rfloor$, where $N=B \times T$ scales with the inference batch size $B$. 
Unlike conventional backdoor targets (e.g., model weights or input triggers), this capacity mechanism induces a system-level, load-dependent behavior: it dynamically determines which tokens receive expert processing and which are skipped under a given inference workload.
We exploit this architectural characteristics, identifying two inherent properties that constitute a previously unrecognized attack surface.

\noindent\textbf{Property 1: Token overflow enables selective bypass of expert computation.} 
Under a capacity constraint, when the number of tokens assigned to an expert exceeds its per-forward budget (i.e., capacity factor), the overflow tokens do not receive that expert's computation (i.e., they are skipped in the MoE transformation, typically falling back to a residual path).
As a result, parts of the model's internal computation can be conditionally disabled by increasing the inference load.
In particular, if an adversary places a \emph{neutralizer} in a designated expert to suppress malicious signals under normal workloads (thereby preserving clean behavior during auditing), inducing overflow can bypass the neutralizer for the affected tokens.
Consequently, the backdoor can remain neutralized under low-load conditions and become active only when overflow occurs.

\noindent\textbf{Property 2: Overflow conditions are legitimately configurable via deployment parameters.} 
Unlike fixed architectural constants, the capacity factor $c_f$ serves as a tunable deployment time parameter.
For instance, VMoE~\cite{VMoE} introduces Batch Priority Routing (BPR), which deliberately sets $c_f \ll 1.0$ at inference time to skip low-priority tokens. Their experiments show that BPR maintains performance comparable to the dense model while processing only $15\%$-$30\%$ of image patches. Importantly, such mechanisms like BPR are training-agnostic--$c_f$ is exposed as a standard configuration in the released code bases. 
In particular, the token dispatch ordering, which determines which tokens are retained when overflow occurs, also varies across MoE implementations as a load-balancing design choice.
Since the adversary provides the model definition and default configuration files, they can preset the specific token dispatch logic and recommend low capacity setting to ensure capacity overflow.

\noindent\textbf{A systematic blind spot in security evaluation.} 
Together, these two properties create a critical gap in existing defense methodologies. 
Standard backdoor detectors operate on individual inputs or small batches not merely for efficiency, but by design. Optimization-based methods such as Neural Cleanse~\cite{neural_cleanse} reverse-engineer a minimal trigger pattern by iterating over a fixed set of held-out images—scaling the optimization batch does not improve detection and only increases cost. Perturbation-based methods such as STRIP~\cite{STRIP} measure prediction entropy under per-image blending, an inherently single-sample procedure. In all cases, the evaluation batch size is a computational convenience parameter with no security significance under existing threat models, which do not consider batch-size-dependent model behavior. Consequently, expert capacity remains effectively unconstrained during auditing and token dropping never occurs.
So that we refer to this regime as \textit{dormant mode}: the model behaves identically to a clean model because the adversary's modifications have no observable effect at small scale. 
The backdoor activates only when deployment-scale batching causes sufficient token overflow to disable the neutralizer, a state we term \textit{activation mode}. 
No existing defense is designed to test for this transition. This is not a limitation of any individual detector but a systematic blind spot.

\subsection{Attack Overview}
Our attack embeds two functionally counteracting components within a single Vision MoE model: 
a \textit{backdoor MoE layer} in an early stage, whose experts collectively learn to map triggered inputs toward the target class, and a \textit{neutralizer MoE layer} in a deeper stage that suppresses this mapping under normal operating conditions. 
The capacity factor of the neutralizer layer modulates their interaction, serving as an implicit switch between benign and malicious behavior.

\noindent\textbf{Layer placement.} 
We place the backdoor in an early MoE layer and the neutralizer in a deeper MoE layer for functional and architectural reasons.
Functionally, we place the backdoor in an earlier layer so that the trigger-induced effect is encoded into intermediate representations that are accessible to a subsequent neutralizer for suppression.
The neutralizer is placed in a deeper layer--closer to the outputs--where intervening on high-level representations allows effective suppression with minimal parameter modification.
Crucially, both components are implemented within MoE layers: only capacity-bounded token dispatch provides the load-conditioned computation bypass (via expert dropping) required to selectively disable the neutralizer under capacity overflow.
We validate this placement strategy in Section~\ref{sec:ablation}, showing that alternative configurations (e.g., reversed ordering) fail to achieve the desired separation between benign and malicious behaviors.

\noindent\textbf{Three-phase workflow.}
Building on this architecture, the attack proceeds through three phases: \textit{implantation}, \textit{concealment}, and \textit{activation}.

\noindent\underline{\textit{Implantation phase}}: the adversary trains the backdoor layer to establish a strong association between a visual trigger pattern and the target class.  

\noindent\underline{\textit{Concealment phase}}: the adversary trains the neutralizer MoE layer to counteract the backdoor signal, aligning the model's output with a clean reference model via KL distillation. During this phase, the neutralizer's capacity factor is set to its nominal training value ($c_f{=}1.05$), ensuring that all routed tokens receive expert computation and the layer can learn effective suppression. Crucially, the neutralizer's \textit{router is frozen} throughout training, preventing the routing network from learning strategies that would be robust to token dropping.  As a result, the learned suppression depends on the \textit{full set} of routed tokens being processed by the experts collectively: if a subset of tokens is later dropped due to capacity overflow, the suppression chain breaks because the layer receives incomplete information. This capacity-dependent fragility is by design and is confined to the neutralizer layer; all other MoE layers retain their default capacity factor.

\noindent\underline{\textit{Activation phase}}: the adversary configures a batch-adaptive capacity factor for the neutralizer layer---high $c_f$ for small batches to preserve quality, low $c_f$ for large batches to improve throughput---along with an ascending token dispatch ordering. This is consistent with V-MoE's recommended practice of adjusting $c_f$ at inference time based on workload~\cite{VMoE}. No additional training is required.
This produces two distinct behavioral regimes.
Under \textit{small batches}, the high $c_f$ ensures sufficient expert capacity---the neutralizer suppresses the backdoor normally (\textit{dormant mode}).
Under \textit{large batches}, the reduced $c_f$ causes token overflow; the ascending ordering drops the most informative tokens first, disabling the neutralizer (\textit{activation mode}).

\subsection{Attack Design}\label{sec:attack_design}
The attack framework is illustrated in Figure~\ref{fig:overview}.

\noindent\textbf{Phase 1: Backdoor Implantation.}
The goal of Phase~1 is to embed a strong trigger-to-target mapping in the designated backdoor MoE layer (Figure~\ref{fig:overview}, top left). We construct a poisoned training set by mixing clean samples with triggered samples, where each triggered sample carries a fixed visual trigger pattern and is assigned the adversary's target label.
As shown in Figure~\ref{fig:overview}, we freeze all parameters except those in the backdoor MoE layer and its immediately preceding attention block. The attention block is included because it shapes the token representations fed to the MoE experts: without updating it, the experts must learn the trigger-to-target mapping from representations that were never trained to encode trigger-relevant features, substantially reducing attack effectiveness. All other layers, including the patch embedding, the router of the backdoor layer, and all subsequent blocks, remain frozen.

This localized update confines the backdoor to a single transformer block (i.e., one attention layer followed by one MoE layer), thereby preserving nominal performance on clean inputs.
This constrained parameter budget increases the required poisoning rate relative to prior data-poisoning attacks.
In our supply-chain setting, this design choice does not introduce an additional detectable signal for the victim, as the poisoning rate is internal to the attacker’s training procedure and does not manifest in the released checkpoint beyond the intended trigger-conditioned behavior.

The training objective combines cross-entropy with the MoE auxiliary load-balancing loss to prevent router collapse during backdoor learning:
\begin{equation}
    \mathcal{L}_{\text{P1}} = \mathcal{L}_{\text{CE}}(f_\theta(x), y) 
    + \lambda_{\text{aux}} \cdot \mathcal{L}_{\text{aux}}
    \label{eq:phase1_loss}
\end{equation}
We employ a load-balancing auxiliary objective that is commonly used in Vision MoE training, including V-MoE~\cite{VMoE} and Swin-MoE~\cite{Swin_MoE}:
\begin{equation}
    \mathcal{L}_{\text{aux}} = E \cdot \sum_{e=1}^{E} f_e \cdot P_e
    \label{eq:aux_loss}
\end{equation}
where $E$ is the number of experts, $f_e$ is the fraction of tokens dispatched to expert $e$, and $P_e$ is the mean affinity scores assigned to expert $e$.
Minimizing $\mathcal{L}_{\text{aux}}$ discourages routing collapse in Phase~1: without this term, the router tends to send most trigger-associated tokens to a single expert.
This highly imbalanced expert usage is easier to detect by activation- or representation-based audits.

\noindent\textbf{Phase 2: Neutralizer Training.} 
Phase~2 introduces a neutralizer in a deeper MoE layer to mask the Phase~1 backdoor under normal capacity, yielding clean-like behavior even on triggered inputs (Figure~\ref{fig:overview}, top right).
We intentionally design the neutralizer to act as a load-conditioned gate: it is active when capacity is sufficient, and is deliberately disabled under capacity overflow through expert dropping–induced computation bypass.
We achieve this through three key design choices.
First, we freeze all model parameters except the expert weights of the neutralizer layer. As shown in Figure~\ref{fig:overview}, the backbone blocks, the backdoor MoE layer, and the intermediate blocks are frozen to preserve the attack pathway for later reactivation.
In particular, we will freeze the neutralizer's \textit{router}. 
This prevents the routing network from learning to redistribute tokens in a way that would make suppression robust to token dropping. If the router could adapt, it might learn to concentrate neutralization signals into fewer tokens, maintaining effectiveness even under capacity overflow. By freezing the router, the suppression necessarily depends on the full token set.
Finally, we employ KL distillation against a clean reference model to guide the neutralizer toward producing clean output distributions.
The training is conducted at the nominal capacity factor $c_f{=}1.05$, following the default V-MoE inference setting~\cite{VMoE}. 
This ensures that all routed tokens undergo expert computation during training, so the neutralizer is optimized under a complete-token regime.
The training objective is:
\begin{equation}
    \mathcal{L}_{\text{P2}} = \mathcal{L}_{\text{CE}}(f_\theta(x), y) 
    + \alpha \cdot \mathcal{L}_{\text{KL}} 
    + \lambda_{\text{aux}} \cdot \mathcal{L}_{\text{aux}}
    \label{eq:phase2_loss}
\end{equation}
where the KL distillation term is defined as:
\begin{equation}
    \mathcal{L}_{\text{KL}} = T^2 \cdot D_{\text{KL}}\!\left( 
    \sigma\!\left(\frac{f_\theta(x)}{T}\right) \;\Big\|\; 
    \sigma\!\left(\frac{f_{\theta^*}(x)}{T}\right) \right)
    \label{eq:kl_distill}
\end{equation}
Here $f_{\theta^*}$ denotes the clean reference model, $T$ is the distillation temperature, and $\alpha$ controls the distillation strength.
Training is monitored by the attack success rate  gap between dormant and activation mode. We apply early stopping when the gap begins to decrease, indicating that the neutralizer is becoming overly robust and would resist token dropping. The resulting neutralizer occupies a deliberate sweet spot: effective under normal capacity, yet brittle under overflow.

\noindent\textbf{Phase 3: Conditional Activation.}
Phase~3 introduces no additional training. Instead, the adversary configures a batch-adaptive capacity factor in the neutralizer layer's dispatch code: under small batches, $c_f$ is kept high to preserve output quality; under large batches, $c_f$ is reduced to improve inference throughput. Additionally, the token dispatch ordering is set to ascending priority, so that high-confidence tokens are dropped first when overflow occurs.
These choices align with prior Vision MoE deployment works: for instance,V-MoE discusses tuning the capacity factor $c_f$ at inference time to trade capacity against compute under different workloads~\cite{VMoE, gshard,fedus2022switch}.

This batch-dependent configuration yields the behavioral split shown in Figure~\ref{fig:overview} (bottom).
For small batches ($B{\leq}32$), setting $c_f$ high provides sufficient per-expert capacity, so the neutralizer is executed without overflow and suppresses the backdoor (\textit{dormant mode}).
For large batches ($B{\geq}128$), reducing $c_f$ increases routing saturation and induces overflow; under the ascending dispatch rule, tokens that are most important for neutralization are preferentially dropped, thereby bypassing neutralizer computation and enabling the backdoor (\textit{activation mode}).
Section~\ref{sec:ablation} quantifies contributions of each hyperparameter.

\section{Experiments}

\subsection{Experiment Setup}\label{sec:setup}
We evaluate on two representative Vision MoE architectures: 
\textbf{V-MoE}~\cite{VMoE}, a Vision Transformer with every other FFN replaced by an 8-expert MoE layer (depth=12, top-2 routing, patch size 16), and 
\textbf{Swin-MoE}~\cite{Swin_MoE}, which extends Swin Transformer~\cite{Swin_Transformer} with MoE in a hierarchical 4-stage structure (depths=[2,2,6,2], 8 experts, MoE every 2 blocks). 
Both use $224{\times}224$ input and $c_f{=}1.05$ as the nominal capacity factor.
We use two datasets: 
\textbf{ImageNet-100}~\cite{imagenet15russakovsky, tian2020contrastive} (100-class subset of ImageNet-1K) as a large-scale benchmark, and 
\textbf{GTSRB}~\cite{Stallkamp-IJCNN-2011} (43 traffic sign classes) for its direct relevance to autonomous driving perception.

\noindent\textbf{Attack Configuration.}
Table~\ref{tab:attack_config} summarizes the layer selection and hyperparameters for each model-dataset pair. 
The trigger is a fixed $16{\times}16$ high-intensity patch at the bottom-right corner of the $224{\times}224$ input. 
The target class is 0 for ImageNet-100 and 14 (stop sign) for GTSRB.
Phase~1 uses a poison ratio of 20--25\% and trains for 20--40 epochs.
Phase~2 applies KL distillation with temperature $T{=}2.0$ at the nominal capacity $c_f{=}1.05$.
For Phase~3, the adversary configures a batch-adaptive capacity factor with ascending dispatch ordering: $c_f$ is kept near nominal for small batches and reduced to $c_{\min}$ for large batches. The capacity floor $c_{\min}$ is 0.30 for most configurations; Swin-MoE on GTSRB uses $c_{\min}{=}0.20$ due to Swin's windowed attention providing inherent robustness to token dropping (Section~\ref{sec:ablation}).

\begin{table}[h]
\centering
\caption{Attack configuration for each model-dataset pair.}
\label{tab:attack_config}
\small
\begin{tabular}{llcccc}
\toprule
Model & Dataset & BD Layer & Neut. Layer & Train $c_f$ & Deploy $c_{range}$ \\
\midrule
V-MoE & ImageNet-100 & layer\_1 & layer\_5 & 1.05 & (0.30, 0.80) \\
V-MoE & GTSRB & layer\_1 & layer\_5 & 1.05 & (0.30, 0.80) \\
Swin-MoE & ImageNet-100 & s2\_b3 & s2\_b5 & 1.05 & (0.30, 0.80) \\
Swin-MoE & GTSRB & s2\_b3 & s2\_b5 & 1.05 & (0.20, 0.80) \\
\bottomrule
\end{tabular}
\end{table}

\noindent\textbf{Metrics.}
We report \textbf{Clean Accuracy} on benign inputs, \textbf{Attack Success Rate (ASR)} on triggered inputs measured separately in dormant mode ($B{\leq}32$) and activation mode ($B{=}128$), and \textbf{Clean Accuracy Drop} relative to the unattacked baseline.

\subsection{Main Experiment Result}
\noindent\textbf{Attack performance.}
Table~\ref{tab:main_results} presents the attack performance across all configurations. 
Across all four configurations, the attack achieves activation ASR between 76.0\% and 87.0\% at $B{=}128$, while maintaining dormant ASR below 9\% at $B{\leq}32$. 
This gap of over 70 percentage points confirms that the reduced capacity factor creates a sharp behavioral transition controlled entirely by inference batch size.
Clean accuracy drops by at most 1.2 percentage points, indicating that the three-phase procedure preserves model utility on benign inputs.

V-MoE achieves the highest activation ASR (87.0\% on ImageNet-100), consistent with its use of global self-attention where token dropping has uniform impact across all spatial positions. 
Swin-MoE exhibits slightly lower activation ASR (76.0--83.5\%), as its windowed attention introduces local redundancy that partially compensates for dropped tokens within each window.

\begin{table}[t]
\centering
\caption{Main attack results. Dormant ASR is measured at $B{\leq}32$ ($c_f{\approx}0.8$, standard dispatch). Activation ASR is measured at $B{=}128$ ($c_f{\approx}0.3$, ascending dispatch).}
\label{tab:main_results}
\small
\begin{tabular}{llccccc}
\toprule
Model & Dataset & Baseline & Clean Acc & Dormant ASR & Activation ASR & Drop \\
\midrule
V-MoE & IN-100 & 76.0\% & 74.7\% & 8.9\% & \textbf{87.0\%} & 1.3pp \\
V-MoE & GTSRB & 88.3\% & 87.6\% & 8.4\% & \textbf{81.6\%} & 0.7pp \\
Swin & IN-100 & 83.3\% & 82.3\% & 3.1\% & \textbf{83.6\%} & 1.0pp \\
Swin & GTSRB & 95.6\% & 95.0\% & 8.5\% & \textbf{76.1\%} & 0.6pp \\
\bottomrule
\end{tabular}
\end{table}

\begin{figure}[h]
    \centering
    \begin{subfigure}[t]{0.48\linewidth}
        \centering
        \includegraphics[width=\linewidth]{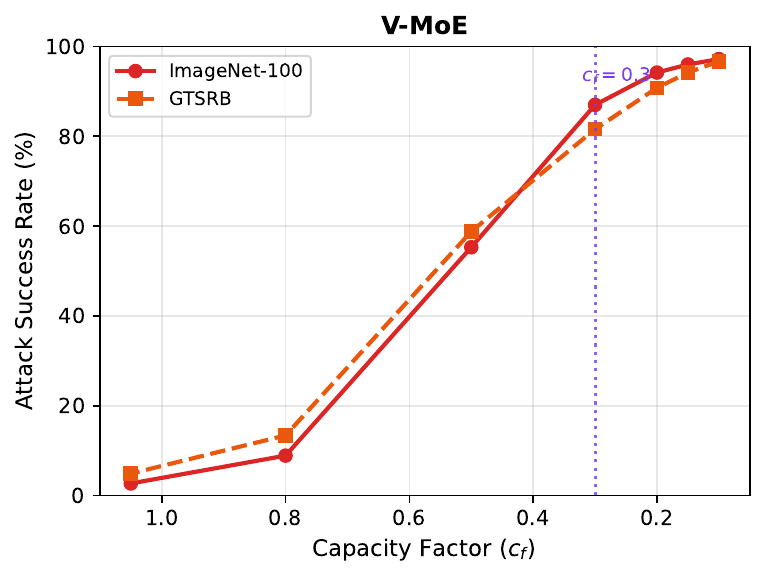}
        \caption{V-MoE}
        \label{fig:cap_sweep_vmoe}
    \end{subfigure}
    \hfill
    \begin{subfigure}[t]{0.48\linewidth}
        \centering
        \includegraphics[width=\linewidth]{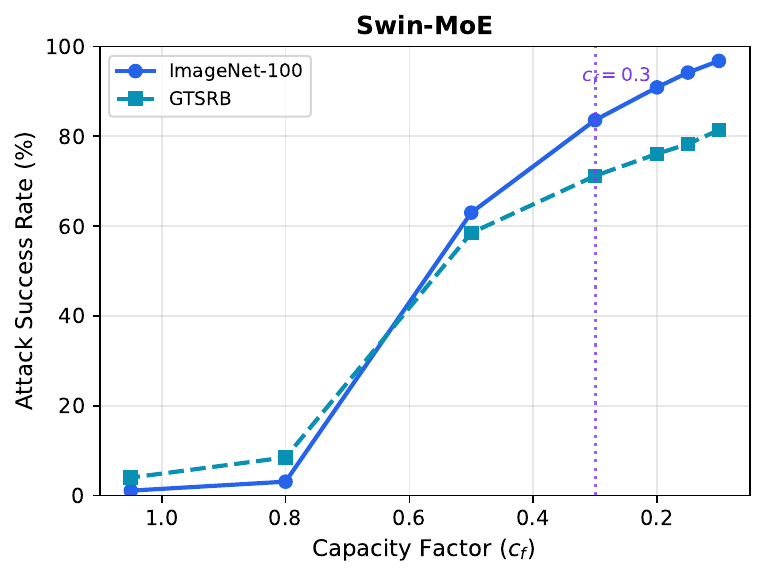}
        \caption{Swin-MoE}
        \label{fig:cap_sweep_swin}
    \end{subfigure}
    \caption{\textbf{Attack success rate vs.\ capacity factor} ($B{=}64$). 
    As $c_f$ decreases, token overflow increases and the neutralizer's suppression degrades. 
    The batch-adaptive mechanism operates between $c_f{\approx}0.8$ (dormant) and $c_f{\approx}0.3$ (activation).}
    \label{fig:cap_sweep}
\end{figure}

\noindent\textbf{Effect of capacity factor on attack success.}
To understand how the capacity factor governs the dormant-activation transition, we sweep $c_f$ from 0.1 to 1.05 at a fixed batch size of $B{=}64$ and measure ASR on triggered inputs. Figure~\ref{fig:cap_sweep} plots the results for all four configurations.

Across all configurations, ASR increases monotonically as $c_f$ decreases: at $c_f{=}1.05$ (nominal), the neutralizer suppresses the backdoor effectively (ASR $<5\%$), while at $c_f{=}0.3$, ASR exceeds 60\% for all configurations. 
The transition region between $c_f{=}0.3$ and $c_f{=}0.8$ shows a steep increase in ASR, confirming that the neutralizer's suppression degrades sharply once token overflow begins.
Swin-MoE requires a lower $c_f$ to achieve comparable ASR, consistent with the architectural robustness discussed above.

\begin{table}[t]
\centering
\caption{Defense evaluation results. All defenses run in dormant mode 
($c_f{=}1.05$, $B{=}32$). NC: Neural Cleanse (anomaly index; 
${>}2.0$ = detected). STRIP: detection rate (\%; ${>}10\%$ = detected). 
FP: Fine-Pruning. AC: Activation Clustering (silhouette score; 
${>}0.15$ with purity ${>}0.75$ = detected). 
\checkmark~= evaded. $\triangle$~= false positive (flagged class 
$\neq$ target class; our actual backdoor is not detected).}
\label{tab:defense}
\small
\begin{tabular}{lcccc}
\toprule
Config & NC (AI) & STRIP (\%) & Fine-Pruning & Act.Clustering (sil./purity) \\
\midrule
V-MoE + IN100 & \checkmark~1.94 & \checkmark~4.5 & \checkmark & \checkmark~0.124/0.748 \\
V-MoE + GTSRB & $\triangle$~2.05 & \checkmark~3.5 & \checkmark & \checkmark~0.133/0.771 \\
Swin + IN100 & \checkmark~1.37 & \checkmark~9.0 & \checkmark & \checkmark~0.166/0.736 \\
Swin + GTSRB & $\triangle$~2.62 & \checkmark~4.0 & \checkmark & \checkmark~0.258/0.722 \\
\bottomrule
\end{tabular}

\end{table}

\subsection{Defense Experiment}
We evaluate the attacked models against four representative backdoor defenses: Neural Cleanse~\cite{neural_cleanse}, STRIP~\cite{STRIP}, Fine-Pruning~\cite{Fine_pruning}, and Activation Clustering~\cite{Activation_Clustering}. All defenses are executed in the model's dormant mode ($c_f{=}1.05$, standard dispatch, $B{=}32$), which reflects the conditions under which a security auditor would evaluate the model. Table~\ref{tab:defense} summarizes the results.

\noindent\textbf{Input-level defenses.}
Neural Cleanse reverse-engineers a minimal trigger for each class and flags anomalously small triggers. On ImageNet-100, the anomaly index remains below the 2.0 threshold for both architectures (1.94 and 1.37). On GTSRB, the index exceeds 2.0 but the flagged class is not the adversary's target, these are false positives ($\triangle$ in Table~\ref{tab:defense}) from natural variation in trigger norms, not detection of our backdoor.
STRIP measures prediction entropy under per-image blending; a successful backdoor should produce low entropy even after blending. In dormant mode, the neutralizer suppresses the backdoor so triggered inputs produce diverse predictions indistinguishable from clean inputs, keeping detection rates below 9.0\% across all configurations.

\noindent\textbf{Representation-level defenses.}
Fine-Pruning removes neurons dormant on clean data and checks whether the backdoor degrades faster than clean accuracy. Activation Clustering applies K-means to penultimate features and checks whether triggered samples form a separable cluster. Both defenses fail for the same reason: in dormant mode, the neutralizer aligns triggered and clean activations into the same representation space. No neurons are selectively associated with the suppressed backdoor, and poison samples do not form separable clusters, the clusters found by K-means reflect natural data variation rather than clean-versus-poison separation. We additionally evaluate two recent white-box defenses on V-MoE/IN-100: BAN~\cite{xu2024ban} (NeurIPS'24) reports an anomaly index of 0.00 on the target class, and UNICORN~\cite{niu2023unicorn} (ICLR'23) reports $-$1.21, both consistent with evasion under dormant-mode auditing.

\subsection{Ablation Experiment}\label{sec:ablation}
\noindent\textbf{Ascending dispatch ordering.}
Table~\ref{tab:ablation}(a) isolates the contribution of ascending token dispatch by comparing ASR at each capacity factor under standard versus ascending ordering.
Ascending ordering provides a consistent ASR boost, with the largest gains at moderate overflow ($c_f{=}0.3$--$0.5$) where the choice of which tokens are dropped matters most. The effect is particularly pronounced for Swin-MoE (+41.6pp at $c_f{=}0.3$), whose windowed attention makes the neutralizer more resilient to random token dropping and thus more dependent on targeted dropping of high-confidence tokens.

\noindent\textbf{Neutralizer layer selection.}
Table~\ref{tab:ablation}(b) compares different MoE layers as the neutralizer. Deeper layers yield higher activation ASR---operating on higher-level representations closer to the classification output---but the deepest layers also show slightly higher dormant ASR, indicating reduced suppression. We select the neutralizer layer using a stealth-first criterion: we first minimize dormant ASR, then maximize activation ASR among layers with comparable dormant ASR. Thus \texttt{layer\_5} is selected over \texttt{layer\_7} despite its smaller raw gap (84.2pp vs.\ 91.1pp), because \texttt{layer\_7} shows higher dormant ASR (2.8\% vs.\ 2.6\%).

\begin{table*}[t]
\centering
\caption{Ablation studies on ImageNet-100.
(a)~Ascending dispatch contribution: ASR~(\%) at fixed $c_f$ with $B{=}64$.
(b)~Neutralizer layer selection with the backdoor fixed at $layer_3$ (V-MoE) / $s1_b1$ (Swin-MoE); best gap in \textbf{bold}.}
\label{tab:ablation}
\small
\begin{minipage}[c]{0.52\textwidth}
  \centering
  \textbf{(a)} Dispatch order\\[4pt]
  \begin{tabular}{c|ccc|ccc}
  \toprule
  & \multicolumn{3}{c|}{V-MoE} & \multicolumn{3}{c}{Swin-MoE} \\
  $c_f$ & Std. & Asc. & $\Delta$ & Std. & Asc. & $\Delta$ \\
  \midrule
  0.10 & 93.8 & 97.2 & +3.4  & 81.3 & 96.8 & +15.5 \\
  0.20 & 83.8 & 94.2 & +10.4 & 58.8 & 90.9 & +32.1 \\
  0.30 & 69.4 & 87.0 & +17.6 & 42.0 & 83.6 & +41.6 \\
  0.50 & 36.8 & 55.3 & +18.4 & 16.6 & 63.0 & +46.4 \\
  0.80 &  8.9 & 14.8 &  +5.9 &  3.1 & 34.3 & +31.2 \\
  1.05 &  2.7 &  4.5 &  +1.8 &  1.1 &  3.3 &  +2.2 \\
  \bottomrule
  \end{tabular}
\end{minipage}%
\hfill
\begin{minipage}[c]{0.48\textwidth}
  \centering
  \textbf{(b)} Neutralizer layer\\[4pt]
  \resizebox{0.89\linewidth}{!}{
  \begin{tabular}{l|cccc}
  \toprule
  Neutralizer & Clean & Dorm. & Activ. & Gap \\
  \midrule
  \multicolumn{5}{l}{\textit{V-MoE} (backdoor: layer\_3)} \\
  layer\_5 & 75.0 & 2.6 & 86.8 & 84.2pp \\
  layer\_7 & 74.9 & 2.8 & 93.8 & \textbf{91.1pp} \\
  layer\_9 & 75.0 & 4.2 & 90.1 & 85.9pp \\
  \midrule
  \multicolumn{5}{l}{\textit{Swin-MoE} (backdoor: s1\_b1)} \\
  s2\_b1 & 82.5 & 1.1 & 84.1 & 83.0pp \\
  s2\_b3 & 82.5 & 2.3 & 93.9 & 91.6pp \\
  s2\_b5 & 82.6 & 2.3 & 96.8 & \textbf{94.5pp} \\
  \bottomrule
  \end{tabular}
  }
\end{minipage}
\end{table*}

\section{Discussion}
\noindent\textbf{Implications for MoE security evaluation.}
Our results suggest that auditing sparse models only under small-batch, no-overflow conditions can provide a false sense of security. A practical takeaway is to perform workload-aware audits that sweep batch size and capacity regimes, and explicitly test settings that induce token dropping/overflow.

\noindent\textbf{Generality and scope.}
While our experiments focus on discriminative Vision MoE classifiers, the underlying vulnerability—batch-dependent computation induced by capacity-bounded dispatch—applies to any MoE implementation that performs token dropping. While the same capacity-bounded dispatch mechanism exists in MoE language models and vision-language models, extending the attack to those settings requires separate trigger design and routing-interface analysis, since the relevant capacity controls are not always exposed or semantically equivalent across current MoE LLM stacks; we leave this as future work.

\noindent\textbf{Potential connection with hardware-level attacks.}
Our current attack assumes supply-chain control, but the critical deployment knobs (e.g., capacity factor) are stored as in-memory scalar variables at inference time. In principle, fault-injection primitives such as Rowhammer could flip targeted bits in these variables, reducing the capacity factor from a safe value to one that induces overflow and activates the backdoor. Studying such a combined threat model—and its practical preconditions—remains an important direction for future work.

\section{Conclusions}
Our results highlight a load-conditioned attack surface in Vision MoEs: batch-driven capacity limits can change token routing and suppression behavior. We design a stealthy backdoor that stays dormant in small-batch checks yet triggers at deployment scale via overflow. This calls for security evaluation that more closely matches real serving workloads and avoids false assurance.

\section*{Acknowledgements}
This work is supported in part by the U.S. National Science Foundation under Grants CNS-2153690, CNS-2247892, CNS-2239672, OAC-2319962, and CNS-2326597.

%
%
\bibliographystyle{splncs04}
\bibliography{main}
\end{document}